\documentclass[final]{cvpr} 

\usepackage{graphicx}
\usepackage{amsmath,amssymb,amsfonts}
\usepackage{multicol} 
\usepackage{caption}
\usepackage{wrapfig} 
\usepackage[pagebackref=true,breaklinks=true,colorlinks,bookmarks=false]{hyperref}
\usepackage{booktabs}
\usepackage{subfigure}
\usepackage{array,multirow}
\usepackage{float}
\usepackage[ruled,vlined]{algorithm2e}
\usepackage{amssymb,amsmath,amsthm}

\def\cvprPaperID{4335} 
\def\confYear{CVPR 2021}

\newtheorem{thm}{Theorem}
\newtheorem{lemma}[thm]{Lemma}

\newcommand\ours{{BinPlay}}

\def\Eparam{\nu}     
\def\Dparam{\upsilon}

\def\natural{\mathbb N}

\def\comment#1{}

\def\eqref#1{(\ref{#1})}
\def\Beq#1\Eeq{\begin{equation}#1\end{equation}}
\def\Beqo#1\Eeqo{\begin{equation*}#1\end{equation*}}
\def\Beqs#1\Eeqs{\begin{align}#1\end{align}}
\def\Beqso#1\Eeqso{\begin{align*}#1\end{align*}}

\begin{document}

\title{\ours{}: A Binary Latent Autoencoder\\for Generative Replay Continual Learning}  

\author{Kamil Deja$^1$, Paweł Wawrzyński$^1$, Daniel Marczak$^1$, Wojciech Masarczyk$^1$, Tomasz Trzciński$^{1,2}$\\
$^1$Warsaw University of Technology $^2$Tooploox\\
{\tt\small \{kamil.deja.dokt,pawel.wawrzynski,daniel.marczak.stud,wojciech.masarczyk.dokt,tomasz.trzcinski\}@pw.edu.pl}
}

\maketitle
              
\begin{abstract}
We introduce a binary latent space autoencoder architecture to rehearse training samples for the continual learning of neural networks. The ability to extend the knowledge of a model with new data without forgetting previously learned samples is a fundamental requirement in continual learning. Existing solutions address it by either replaying past data from memory, which is unsustainable with growing training data, or by reconstructing past samples with generative models that are trained to generalize beyond training data and, hence, miss important details of individual samples.  
In this paper, we take the best of both worlds and introduce a novel generative rehearsal approach called~\ours{}. Its main objective is to find a quality-preserving encoding of past samples into precomputed binary codes living in the autoencoder's binary latent space. 
Since we parametrize the formula for precomputing the codes only on the chronological indices of the training samples, the autoencoder is able to compute the binary embeddings of rehearsed samples on the fly without the need to keep them in memory. 
Evaluation on three benchmark datasets shows up to a twofold accuracy improvement of \ours{} versus competing generative replay methods.
\end{abstract}

\section{Introduction} 

\begin{figure}[b]
\centering
\includegraphics[width=.43\textwidth]{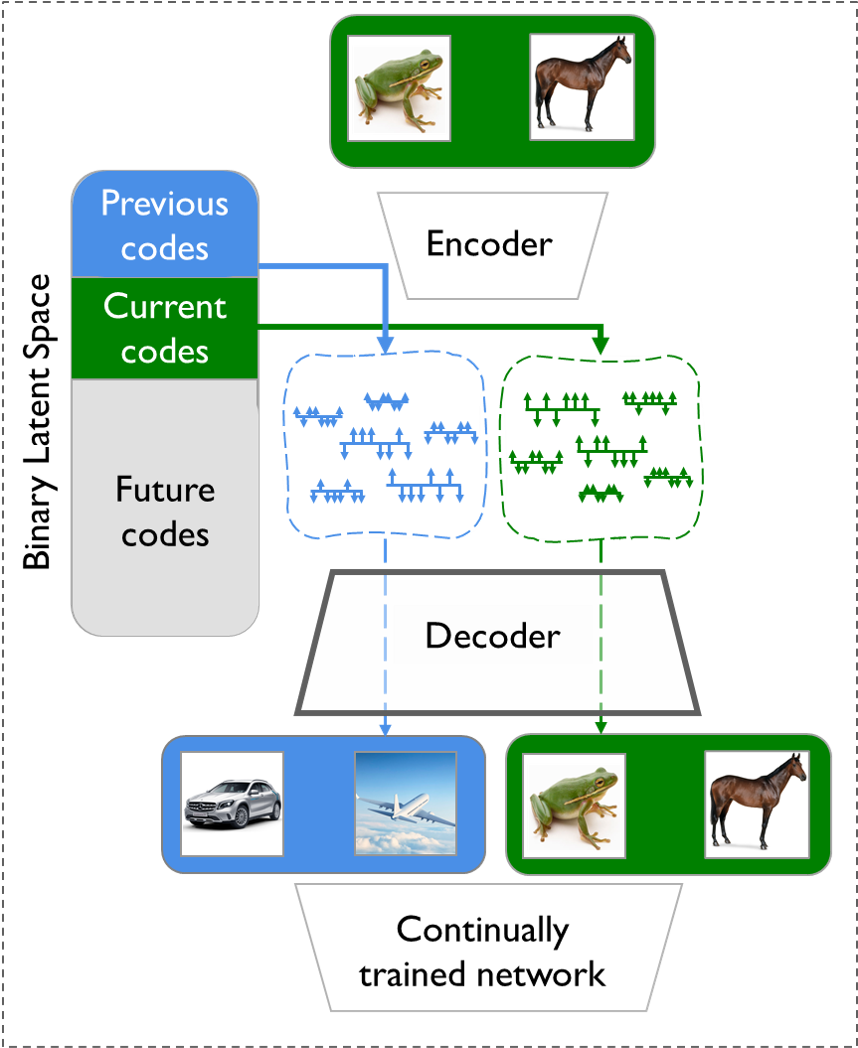}
\caption{Overview of our \ours{} architecture for continual learning. When a new batch of data arrives, we train the encoder to map new images to a set of precomputed binary codes in the latent space and then we train the decoder to reconstruct them back to their originals. Since we compute the binary codes using only the indices of training data samples, we know which part of the latent space is populated with the codes of previously seen data. We can also re-compute those codes on the fly, using the indices, and sample them uniformly to rehearse past samples. This way, our \ours{} model can encode information about the past data in the binary latent space of the autoencoder, without the need to store individual images or their codes in memory.}
\label{fig:teaser} 
\end{figure} 

While mammal brains are capable of acquiring new knowledge without forgetting the skills learned in the past~\cite{1999french}, artificial neural networks fail to copy this behavior. The resulting \emph{catastrophic forgetting} phenomenon~\cite{1999french} manifests itself in the deteriorating performance of a model on a previously learned task upon learning a new one. This is due to the fact that in the training phase, a neural network assumes a stationary data distribution that does not change in time. 
Hence, re-training the network with new input data adjusts its weights to improve the performance on the current objective, disregarding previously seen samples and the associated knowledge~\cite{2017kirkpatrick+many}. The assumption of stationary data distribution often fails in practice, as numerous applications,  including experimental physics~\cite{tilaro2018model}, autonomous driving~\cite{yang2018pixor}, robotics~\cite{kehoe2015survey}, and recommendation engines~\cite{he17}, observe ever-growing datasets with changing data characteristics.
 
 A domain of machine learning that attempts to address these requirements is referred to as \emph{continual learning}. Methods of continual learning proposed in the literature include regularization of neural network weights~\cite{2017zenke+2,2017kirkpatrick+many}, adjustments in the structure of a network to the next batches of data\footnote{In continual learning, a new portion of data fed to the model is often referred to as a {\it task}, but to avoid confusion with multi-task learning we use a term {\it batch} of data in this paper.}~\cite{2018xu+1,2017yoon+3,2016rusu+7,2019golkar+2,2019cheung+4,2020wen+2}, or replaying examples from previously seen batches while training the network with new data~\cite{rebuffi2017icarl, lopezpaz2017gradient}, also known as rehearsal training.
 As storing past data requires a growing buffer, recent methods for replaying past samples leverage generative neural architectures~\cite{2014goodfellow+7,kingma2014autoencoding} to reconstruct previously seen data points from various probability distributions~\cite{van2018generative,2019xiang+3,2020mundt+4}.
 
 These models, referred to as generative rehearsal methods, theoretically allow infinite continual learning without inflating the buffer of past examples. However, their performance is upper-bounded by the quality of the training data created with generative models. Since the most frequently used generative models, such as Generative Adversarial Networks (GAN)~\cite{2014goodfellow+7} and Variational Autoencoders (VAE)~\cite{kingma2014autoencoding}, are trained to learn a continuous function that describes the entire dataset distribution, they often yield visually plausible %
 yet low-quality results. For more challenging datasets, generative models often output blurry images that miss high-frequency details~\cite{lesort2019generative,aljundi2019online}. When used as samples for the rehearsal procedure, such examples lead to the deteriorated performance of neural network models~\cite{lesort2019generative}.

In this paper, we address these shortcomings by introducing an autoencoder-based architecture named \ours{}\footnote{We make the code available at \url{https://github.com/danielm1405/BinPlay}} that is able to regenerate the samples seen in the previous batches out of a set of precomputed binary codes. Inspired by generative hashing approaches~\cite{carreira15,mena19,zamorski20}, we achieve that goal by designing a binary latent space autoencoder that learns the most quality-preserving mapping between current batch images and a set of binary vectors living in the autoencoder latent space. We allocate the pool of binary codes for a current batch using only a chronological ordering of training samples, {\it i.e.}, the ordinal number at which samples are presented to the model. Since we know the number of data samples seen in previous batches, we can rehearse past images by first sampling binary codes uniformly from the pool of codes already {\it allocated} and then decoding them with an autoencoder. This way, we do not need to store any binary embeddings for previous batches in memory. 
At the same time, we avoid the problem faced by competing generative rehearsal methods, namely the implicit requirement for the generative model to generalize beyond a set of training samples. 
Fig.~\ref{fig:teaser} illustrates the overview of our method. 

Although our approach can be easily generalized to other computer vision applications, in this work, we focus on the image classification benchmarks. We evaluate our~\ours{} method against competitive approaches 
on the most challenging \emph{class incremental} learning scenario. \ours{} consistently outperforms state-of-the-art methods on three benchmark datasets while providing a comparable or lower model memory footprint. 


To summarize, the contributions of this work are: 
\begin{itemize}
    \item a new approach to continual learning problem called \ours{} that combines advantages of generative and buffer-based rehearsal methods by encoding previously seen samples not in memory but in the binary latent space of an autoencoder,
    \item a novel binary latent space autoencoder architecture which learns a quality-preserving, one-to-one mapping between images and predefined binary codes, 
    \item a binary code assignment method based on the ordinal number of training images, which enables the computation of binary codes corresponding to rehearsal samples on the fly, without the need to store them in memory.
\end{itemize}

The remainder of this paper is organized as follows. Sec.~\ref{sec:related work} overviews the related work. Sec.~\ref{sec:method} introduces our proposed~\ours{} approach and in Sec.~\ref{sec:experiments} we present the experiments confirming validity of our method. We conclude this work in Sec.~\ref{sec:conclusions}.  

\section{Related work} \label{sec:related work} 

Continual or lifelong learning, defined as the ability of a machine learning model to learn new knowledge without forgetting previously learned tasks, attracts increasing attention of the research community~\cite{2020mundt+3}. As simple re-training of neural systems on new data leads to {\it catastrophic forgetting}~\cite{1999french}, there are many alternative approaches in the literature. We can classify them into three categories discussed below. 

\paragraph{Methods based on 
regularization.}\!\!\!\! The idea behind these methods is to train a~model on the incoming batches of new data while at the same time imposing regularization, which enforces preservation of the model performance on the previous data. 
Different methods such as Synaptic Intelligence (SI) \cite{2017zenke+2} and Elastic Weight Consolidation (EWC) \cite{2017kirkpatrick+many} propose different regularization functions. However, their goal is to slow down the weights update procedure for weights that are considered most important for the previous batches of data. Recent experiments indicate that methods based on regularization do not prevent forgetting but usually slow it down~\cite{2020delange+7}. 

\paragraph{Methods based on batch-specific model components.}\!\!\!\!
In these methods, different model versions are built for different batches of data, although sharing some common parts. During inference, the sample is first assigned to the proper batch (or their combination), to target the specific part of the network. In Reinforced Continual Learning (RCL) \cite{2018xu+1}, Dynamically Expandable Networks (DEN) \cite{2017yoon+3}, and Progressive Neural Networks (PNN) \cite{2016rusu+7} new structural elements are added to the model for each new data batch. Alternatively in \cite{2018masse+2,2018mallya+1,2019golkar+2,2018mallya+2} authors introduce different methods for selecting different submodels of the main model for consecutive batches. The methods in this category provide high accuracy in a multitask scenario, when test samples are given with a corresponding batch index~\cite{vandeven2019scenarios}. Otherwise, the index needs to be estimated and current approaches solve that using heuristics, such as the minimization of classification entropy~\cite{2016hendrycks+1,2020wortsman+6}. 


\paragraph{Methods based on replaying.}\!\!\!\! 
This group of methods tries to preserve previous data and use it along with new samples when retraining the model. The first group of works including \cite{2019rolnick+4,aljundi2019online} employ a~memory buffer to store all or possibly most relevant previous data examples. Although this approach directly solves the problem, for proper results, it needs to store at least several examples from each incoming batch. This requirement makes the solution inadequate for the general continual learning problem, in which we would like to retrain the model in the potentially infinite number of batches. 

In \cite{2017shin+3}, the authors propose to replace the buffer with a~generative model based on the Generative Adversarial Networks (GAN) \cite{2014goodfellow+7}. However, since any structure used to compress past data may also suffer from catastrophic forgetting, authors propose a self retraining procedure in which they retrain the generative model with both new data batches and regenerated examples from the previous ones. In our \ours{} method we benefit from this approach while retraining our binary latent autoencoder as described in Sec.~\ref{sec:method}. 

Similarly to~\cite{2017shin+3}, \cite{van2018generative} introduces a generative rehearsal model based on another successful generative method: Variational Autoencoder (VAE) \cite{kingma2014autoencoding}. Additionally, they combine the generative model with the base classifier, which reduces the cost of model retraining. VAE was also used in favor of GANs in ~\cite{2020mundt+4}, where authors successfully incorporated this model to the problem of open datasets. 
An interesting idea that  we can place in between of generative and buffer based rehearsal is presented in~\cite{caccia2020online}, where authors incorporate VQ-VAE~\cite{oord2018neural} architecture to compress the original data examples into a~special representation that requires less memory the original images. In this work we extend this idea by proposing a method that compresses the data examples to the set of specific binary vectors which contrary to VQ-VAE does not have to be stored in the buffer.



Therefore, the method introduced in this paper falls into the category of generative rehearsal. Our approach is conceptually similar to that in \cite{2020mundt+4}. We train an autoencoder with an~increasing set of images to reproduce them for the rehearsal procedure. 

However, all of the above-mentioned 
procedures are based on the generative models that are trained to approximate functions that produce images. 
While doing so with the mixtures of continuous distributions (e.g. gaussian), the resulting images are often averaged across samples and blurred. Continuous distributions come with benefits, allowing not only to reproduce previous examples but also to generate new ones as the interpolation between them~\cite{berthelot2018understanding}
This valuable asset however, contributes little to the problem of rehearsal-based incremental learning. Therefore, in this work we propose to drop it in favor of high quality reconstructions of the original data examples.

\section{Our method} \label{sec:method} 

\begin{figure*}
\centering
\includegraphics[width=0.92\textwidth]{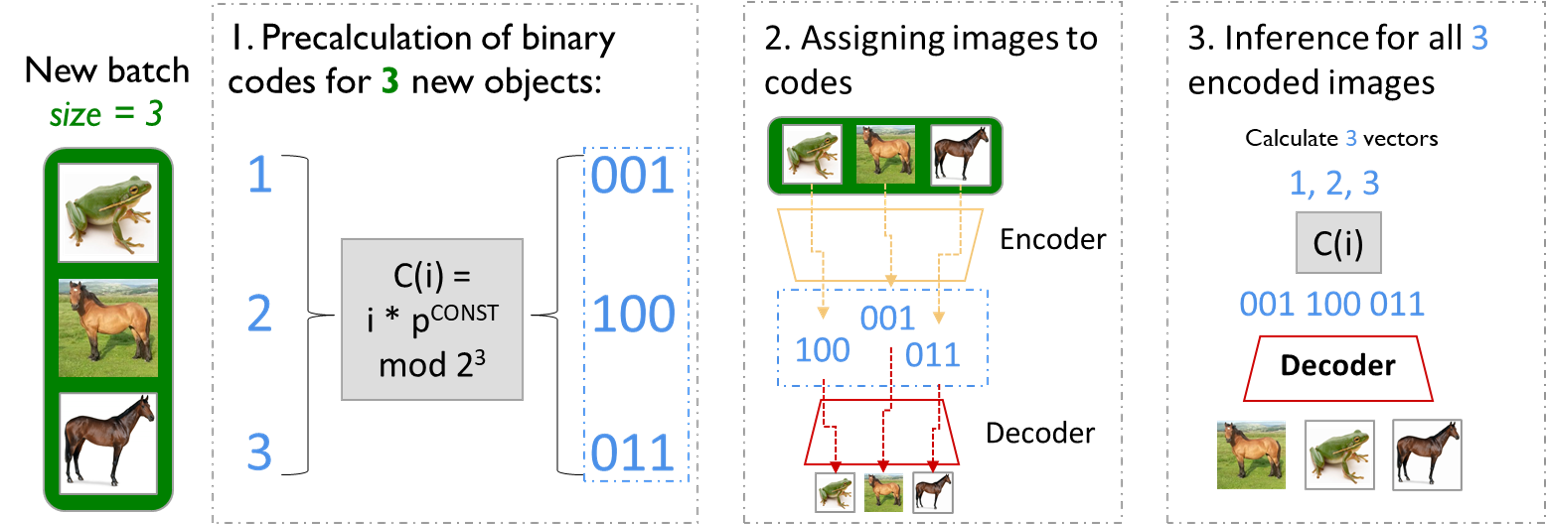}
\caption{Encoding and decoding~\ours{} binary latent space codes. We first create a set of binary codes on the basis of consecutive indices (\textbf{I}). We then train the encoder to find the best mapping between original images and a set of binary codes, and the decoder to reconstruct them back to their originals (\textbf{II}). To generate all previously encoded images, we only need the decoder and the~total number of encoded samples (equal to the last sample's index, if counted from 1), so we can recompute the set of binary codes and process them through the decoder (\textbf{III}).} 
\label{fig:encoding} 
\end{figure*}

The goal of our method is to continually train a base network while rehearsing the data from the past batches to avoid forgetting previously learned knowledge. For the rehearsal, we refer to the binary latent space autoencoder, an auxiliary neural network inspired by the hashing autoencoder approaches~\cite{carreira15,mena19,zamorski20}. The objective of this autoencoder is to find an assignment of the input data to a set of predefined binary codes and then to reconstruct the codes back to the original data. We define the binary codes to live sparsely in the latent space to avoid hashing conflicts and simplify the reconstruction while populating only a limited part of a latent space. This allows us to rehearse previously seen data by uniformly sampling populated latent space and reconstructing the original data from the sampled binary codes. We show an overview of this mechanism in Fig.~\ref{fig:encoding}. The next subsections discuss in more detail the building blocks of our approach, namely the binary latent autoencoder, definition of binary codes, and their assignment to input data.

\subsection{Binary latent autoencoder \label{sec:binary_training}} 

The objective of our autoencoder is to reconstruct a set of input images. Therefore, the encoder needs to learn their mapping to a set of predefined binary codes living in the autoencoder's latent space. The decoder, on the other hand, should be able to reproduce inputs from previous training batches based on the latent binary vectors assigned to them.


In order to train the binary latent autoencoder network to fulfill the above requirements, we follow the procedure introduced in~\cite{2017shin+3} and use both data from a~new batch and the sampled outputs from its current version. Thanks to this approach, we learn mappings and reconstructions for the new samples while preserving the knowledge of the previous ones at the same time.
First, we create a copy, $\Dparam'$, of current decoder weights, $\Dparam$, so that decoder with weights $\Dparam'$ can serve as a~source of previous training pairs for the network.

For the training of our autoencoder, we create a set of training samples by 
uniformly drawing indices $i \in \{1, \dots, K, \dots, N\}$, where $K$ is the first index for the current batch, and $N$ is the total number of images observed so far. 
\begin{itemize}
    \item If $i$ is between $1$ and $K-1$, we recreate an image pair as $\langle c(i), d(c(i);\Dparam')\rangle$, where $c$ is a binary codes generator function and $d$ is our decoder.
    \item If $i$ is between $K$ and $N$, we use a new training pair as $\langle c(i), x_i\rangle$, where $x_i$ is an~original image assigned to the index $i$.
\end{itemize}

With combination of such pairs, we train the original weights of our decoder $\Dparam$ and learn  image reconstructions from their binary codes.
To ensure that the above sampling procedure in the latent space works, we propose an appropriate binary code definition. 

\begin{figure*}[!htb]
\centering
\includegraphics[width=.8\textwidth]{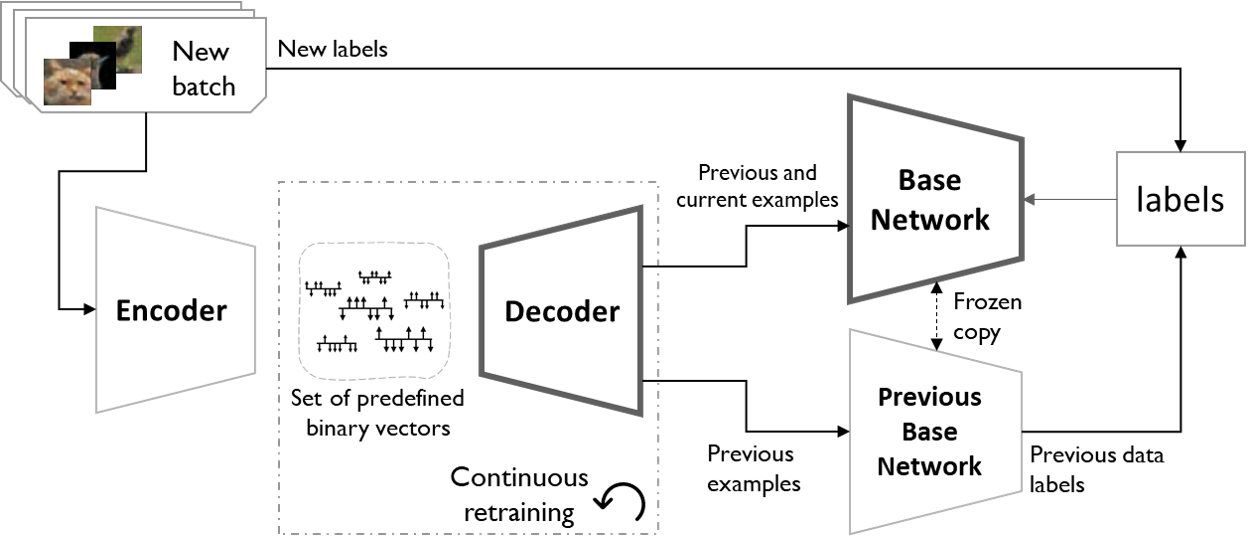}
\caption{The BinPlay architecture consists of two feed-forward networks. The first one -- \textbf{binary latent autoencoder} -- finds the mapping of new training examples to the predefined binary codes and decodes them to the original training examples with the \textbf{decoder}. While training proper reconstruction of the binary codes into new images, we also feed the decoder with the combinations of previously seen codes and their reconstructions in the continuous retraining. We train the \textbf{base network} with the generated examples from the previous data and the current data samples. For the generated examples, we calculate appropriate labels with a frozen copy of the base network. Only the decoder and base network are needed to be stored in the memory buffer, while the predefined binary codes for past data are computed on the fly.} 
\label{fig:architecture} 
\end{figure*}

\subsection{Binary codes definition}

As we want to rehearse previously seen data, we require the set of binary codes within the latent space to be parametrized by an index. Then, in order to produce a set of uniformly sampled~random binary codes that can be used to reconstruct the previously seen data, we only need to sample a set of indices. Last but not least, we require the codes to be composed of a~large number of bits that often change between the subsequent indices. This way, codes become easily recognizable inputs for the decoder network.


To fulfill the above requirements, we follow an observation presented below and proven in the supplementary material:

\begin{lemma} \label{lemat2m} 
Let $p$ be a prime number, $p\neq 2$, and $i \in \{1, \dots, 2^m\}$ be coded as $m$ least significant bits of the integer 
\Beq \label{c(i)}
    i \cdot p^{\lfloor m \ln 2 / \ln p \rfloor}.
\Eeq\\
Then 
\begin{enumerate}
    \item  $2^m/p < p^{\lfloor m \ln 2 / \ln p \rfloor} < 2^m$. 
    \item  The codes for all the numbers $i$ are different. 
\end{enumerate}
\end{lemma}


With this observation, we define a~coding of the sample index, $ c : \natural \mapsto \{0, 1\}^n, $ as follows: 
\begin{enumerate}
    \item The vector $c(i)$ is composed of several subvectors.
    \item For a~subvector of size $m$, we assign the $m$ least significant bits of the binary representation of the formula~\eqref{c(i)} with $i$ as the index, and $p$ as a~prime number different for different subvectors.
\end{enumerate}

As a result of this procedure, the binary codes for the data indices are different and sparsely distributed. Moreover, according to Lemma~\ref{lemat2m}, active bits of the codes calculated with the proposed procedure often change between subsequent codes. Hence they fulfill our requirements. 

In a typical continual learning scenarios, data samples within a training batch share some similarities, as the batches are constructed with the assumption of stationary data distribution. Therefore, we append a batch index transformed using the procedure above as a prefix of the binary codes defined for a given batch.    

\subsection{Binary codes assignment}

Once we define the binary codes for a given data batch, we need a procedure to map the original data into this set of codes. For a decoder to rehearse past samples properly, we need the mapping to provide a single binary code in the latent space for each data sample. We, therefore, propose the following greedy binary code assignment algorithm.

\begin{algorithm}[t!]
\SetStartEndCondition{ }{}{}%
\SetKwProg{Fn}{def}{\string:}{}
\SetKwFunction{Range}{range}
\SetKw{KwTo}{ in }\SetKwFor{For}{for}{\string:}{}%
\SetKwIF{If}{ElseIf}{Else}{if}{:}{elif}{else:}{}%
\SetKwFor{While}{while}{:}{fintq}%
\AlgoDontDisplayBlockMarkers\SetAlgoNoEnd\SetAlgoNoLine%
\caption{\label{alg:code_assignment}Designating regularization losses for binary codes assignment in \ours{}.}
\textbf{Input:} C = set of binary codes, X = set of \\$\quad\;\;$images, $e$ = encoder,  $\Eparam$ = encoders weights \\
put elements of $X$ in a random order \\ 
\For{$x_i \KwTo X$}{
    $z_i = e(x_i;\Eparam)$ \\ 
    $c_i =$ the element of $C$ closest to $z_i$ \\ 
    $C = C \setminus \{c_i\}$ \\ 
    $L_i(z) = \|z-c_i\|^2$
}
\end{algorithm}


For the set of images $X$, we generate a set of binary codes $C$, such that $|X|=|C|\ll 2^n$, where $n$ is a size of the latent space. Then, while training our binary latent autoencoder, we extend the original reconstruction loss with an additional regularization term. To compute this term, we assign a yet unassigned binary code to each sample that lies at the lowest distance to the sample. Then, we calculate the distance between data embedding in the latent space and its associated binary code as a $L_2$ norm. The final regularization is a sum of the distances computed across all of the images in the batch. Since this procedure is highly affected by the ordering of the samples, in each epoch, we shuffle the input images $X$. Alg.~\ref{alg:code_assignment} presents a general overview of the assignment procedure.

\subsection{Trainining} 

With the building blocks of~\ours{} described above, we can now describe how to train our model. Our approach provides a general framework for various continual learning problems, yet here we focus on the incremental image classification as a representative one. Fig.~\ref{fig:architecture} shows the  overview of our method.

On the arrival of a~new batch, we first train the binary latent autoencoder. We train its encoder to assign new examples to their binary codes and the decoder to reconstruct them back to the original form. To retain the knowledge derived from the previous data, we use the training procedure as defined in Sec.~\ref{sec:binary_training}. To enable faster convergence, we start training autoencoder without enforcing the binary regularization of the latent space and turn the binary code assignment of Alg.~\ref{alg:code_assignment} after a couple of {\it warm-up} rounds. This allows the autoencoder to cluster similar images in the latent space before mapping them to the allowed binary codes. Once the encoder converges on the mappings between the images and their binary codes, we save the mappings and omit the encoding loss to focus on the decoder. For training simplicity, we use the binary code values of \{-1,1\}, instead of \{0,1\}, as this facilitates the decoding procedure.

Our base network aims at minimizing the loss averaged over current data as well as a sample of the previous training batches. On the arrival of a new batch, we train the network with the new samples together with the samples regenerated from the binary codes sampled uniformly from the allocated pool in the autoencoder latent space. The new samples are processed through the autoencoder before using them as an input for training so that the base network does not focus on differences between the real and generated data but rather on the general image features that enable accurate classification. For the regenerated past samples, we train the base network to predict the same output distribution values as its copy -- frozen before the training. Consequently, we use soft targets~\cite{hinton2015distilling}, instead of the ground truth labels, to better preserve the knowledge gained in the previous training batches. Tab.~\ref{tab:ablation_studies} shows the results of the ablation study supporting the design choices presented in this section.

\begin{table}
  \centering
  \begin{tabular}{l|l}
    \toprule
    Modification & Results\\
    \midrule
    Reference configuration & 23.8 \\ 
    \begin{tabular}{@{}c@{}}+ processing new samples \\ through autoencoder\end{tabular} & 54.0 \\
    + soft targets & 63.3\\
    \bottomrule
  \end{tabular}
    \caption{Ablation study for the training procedures of the base network. We report the average classification accuracy after the last batch on the CIFAR-10 dataset. As a reference configuration, we take a base network trained on both original (current batch) and generated (previous batches) data samples. Preprocessing incoming new samples with an autoencoder and using soft targets improves the performance of our final continual learning model.}
  \label{tab:ablation_studies}
\end{table}

\section{Experimental study} \label{sec:experiments}

We evaluate \ours{} on three commonly used benchmarks: MNIST~\cite{lecun2010mnist}, Fashion-MNIST~\cite{xiao2017fashionmnist}, and CIFAR-10~\cite{Krizhevsky09learningmultiple}. To simulate the real setting of continual learning, we evaluate our model with the {\it class-incremental} scenario. This means that in each batch of data, we introduce new classes to the scope of our model. In particular, we follow the split-MNIST split-FashionMNIST and split-CIFAR procedure that divides the dataset into 5 separate batches with all training examples of classes: \{0,1\}, \{2,3\}, \{4,5\}, \{6,7\}, \{8,9\}, accordingly.

\paragraph{MNIST and Fashion-MNIST}\!\!\!\!


For both MNIST datasets, we propose the architecture based on a convolutional neural network. For the encoder and the decoder, we use 3 convolutional/transposed convolutional layers and 2 fully connected layers around the latent space of size 200.

Our classifier is based on the LeNet~\cite{lecun1998lenet} architecture with 3 convolutional layers followed by two fully connected ones. We use batch normalization and dropout. Detailed implementation information can be found in the released codebase and in the appendix.


                 
\paragraph{CIFAR-10}\!\!\!\!

For the CIFAR-10 dataset, we employ similar network architectures as for MNIST and Fashion-MNIST but with a greater number of filters. Additionally, to simplify the encoding procedure for more complex images, we extend the binary latent size to 1000.


For the classifier, we only adjust the previous LeNet architecture to the CIFAR-10 image size, which is 32×32×3 pixels.

\begin{table*}[htb!]
  \centering
  \begin{tabular}{c|l||ccc}
    \toprule
    Data & Model & MNIST & Fashion & CIFAR--10\\
    storage&& & MNIST & \\ 
    \midrule
    \multirow{5}{*}{\rotatebox[origin=c]{90}{\parbox[t]{1cm}{\centering Memory buffer}}} & GEM~\cite{lopezpaz2017gradient} & 86.3 $\pm$ 1.4 & 70.3 $\pm$ 0.7 & 17.5 $\pm$ 1.6\\
    &iCARL~\cite{rebuffi2017icarl} & 71.7 $\pm$ 0.5 & 67.7 $\pm$ 0.4 & 32.4 $\pm$ 2.1\\
    &ER~\cite{2019chaudry+6} & 84.5 $\pm$ 1.6 & 70.2 $\pm$ 1.7 & 38.5 $\pm$ 1.7 \\
    &ER--MIR~\cite{aljundi2019online} & 87.6 $\pm$ 0.7 & 69.7 $\pm$ 2.7 & 47.6 $\pm$ 1.1\\
    \midrule
    Both & AQM~\cite{caccia2020online} & 93.6 $\pm$ 0.7 & 67.4 $\pm$ 0.3 & 51.4 $\pm$ 2.2 \\
    \midrule
    \multirow{8}{*}{\rotatebox[origin=c]{90}{\parbox[t]{1cm}{\centering Generative replay}}} & GEN-MIR~\cite{aljundi2019online} & 86.6 $\pm$ 0.3 & 52.4 $\pm$ 1.5 & 18.8 $\pm$ 0.9 \\
    &OCDVAE~\cite{2020mundt+4} & 93.2 $\pm$ 3.7 & 69.9 $\pm$ 1.7 & 21.6 \\
    &GR~\cite{2017shin+3} & 92.5 $\pm$ 0.5 & 68.0 $\pm$ 0.9 & 27.3 $\pm$ 1.3 \\
    &GR+distill~\cite{vandeven2019scenarios} & 95.6 $\pm$ 0.2 & 78.1 $\pm$ 1.0 & 28.4 $\pm$ 0.3\\
    &RTF~\cite{van2018generative} & 95.1 $\pm$ 0.3 & 75.2 $\pm$ 0.8 & 28.7 $\pm$ 0.2 \\
    &\textbf{\ours{} (ours)}& \textbf{97.2 $\pm$ 0.6} & \textbf{81.4 $\pm$ 0.9} & \textbf{63.3 $\pm$ 1.4}\\
    \bottomrule
  \end{tabular}
    \caption{Average accuracy after the final batch in the class incremental scenario (in \% $\pm$ SEM). We report the results for memory-buffer based (top) and generative replay (bottom) methods on MNIST, Fashion MNIST, and CIFAR-10. Our approach clearly outperforms competitive approaches on all three benchmarks.}
  \label{tab:results_acc}
\end{table*}

\subsection{Results} 

\begin{figure}
\centering
\subfigure[]{\includegraphics[height=4.35cm]{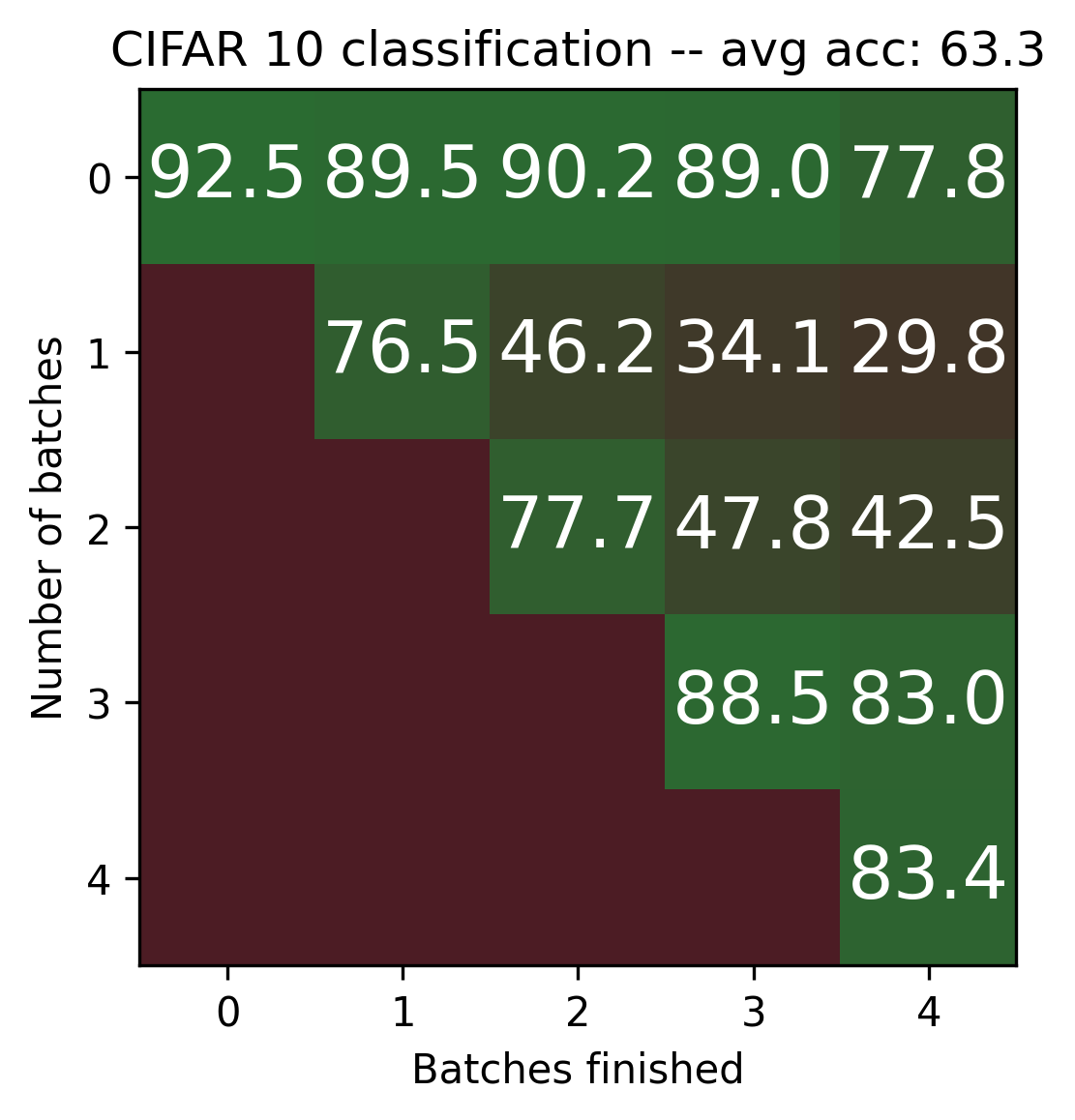}}
\subfigure[]{\includegraphics[height=4.35cm]{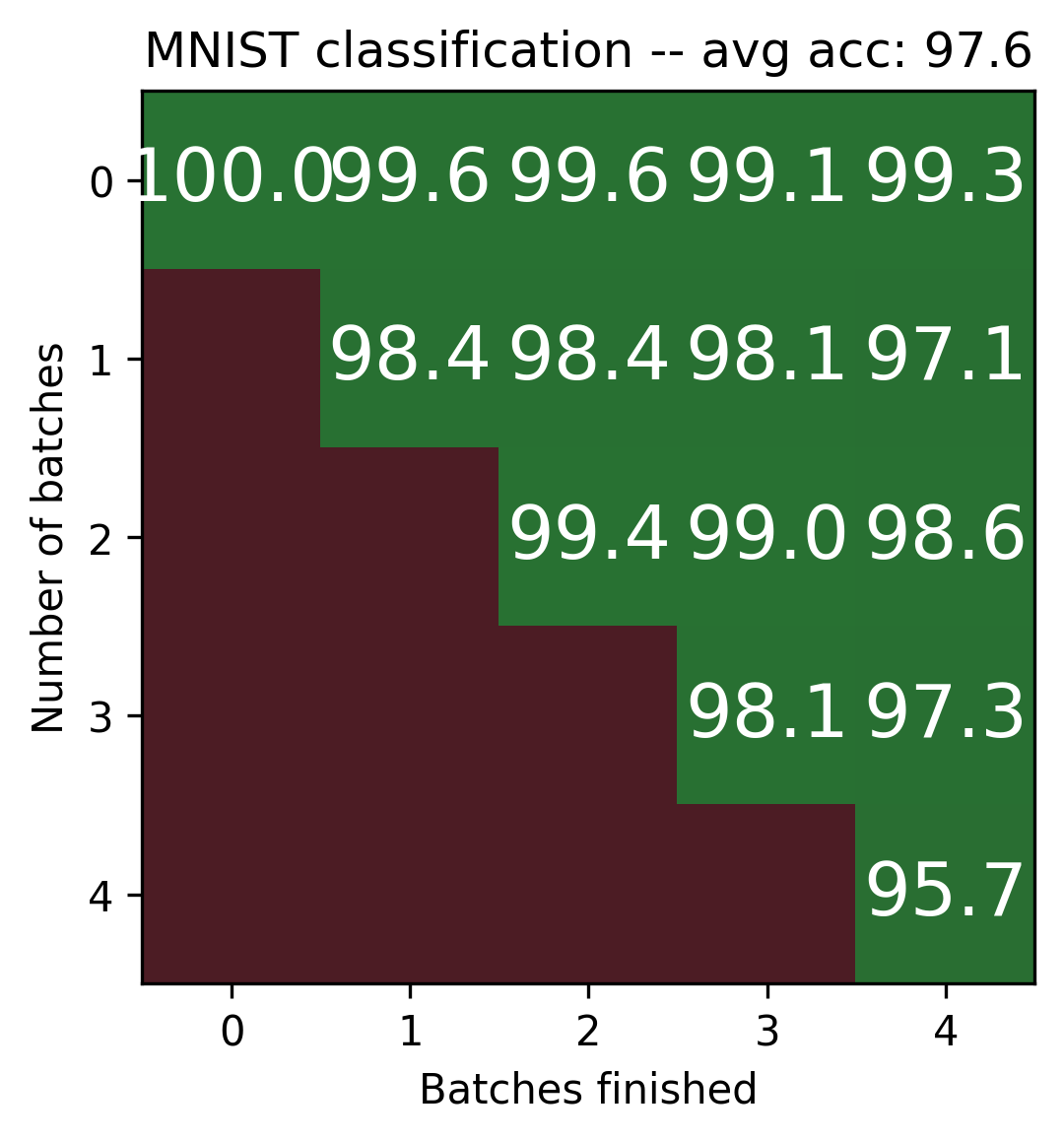}}

\caption{Visualization of the classification accuracy observed on the consecutive batches of the data. For CIFAR-10 (left), the accuracy of our model deteriorates steadily, and it is related to the increasing complexity of the general classification problem rather than the batch index. For simple datasets, such as MNIST (right), we can observe steady performance through all data batches.
}
\label{fig:classification_acc} 
\end{figure}

We compare the accuracy of our models' predictions. For that purpose, we run the experiment through all 5 batches of data to report the average accuracy on the whole test-set after the final one. For the results of competing methods, we report the results from the original works. However, for several techniques, authors openly state that their methods do not scale well for more complex datasets such as CIFAR-10~\cite{aljundi2019online,lesort2019generative}. 
Hence, in case of missing evaluations on any benchmarks, we ran the experiments with the code provided by the authors and report the results averaged across 3 runs. As presented in Tab.~\ref{tab:results_acc}, those experiments demonstrate that our solution clearly outperforms other generative replay approaches on all benchmark datasets. Additionally, in Fig.~\ref{fig:classification_acc} we show that our \ours{} is only slightly affected by the number of consecutive batches and the deterioration of the results is rather the effect of the increasing complexity of the problem.

\begin{figure}
\centering
\includegraphics[width=0.47\textwidth]{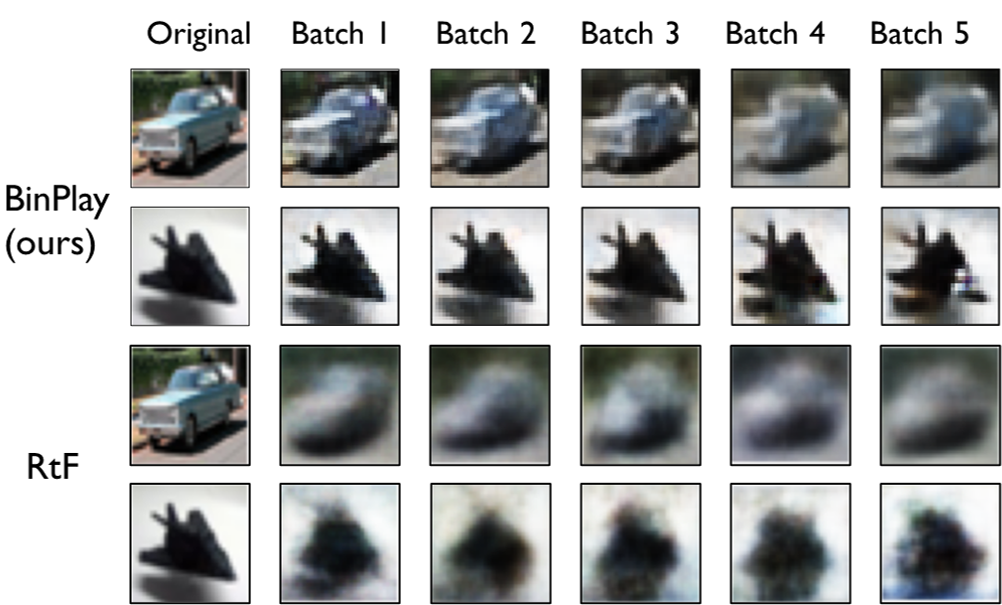}
\caption{Examples of the reconstructions (\ours{}) and generations (RtF~\cite{van2018generative}) of the images from the CIFAR-10 dataset. Our method is able to recreate high-quality images even after four following batches of data. On the contrary to the current state-of-the-art generative replay method (RtF), images are sharp and allow the classifier to rehearse also high-frequency features from regenerated images.} 
\label{fig:generations} 
\end{figure}

This is mainly thanks to the fact that our model can recreate sharp images with high-quality visual features. These are a valuable source of previous examples for the base classifier. To support this claim in Fig.~\ref{fig:generations}, we show that our method is able to recreate rich images even after four batches of not related data examples. This is contrary to the other generative rehearsal procedures such as RtF~\cite{van2018generative} that produce blurry images already from the start.

We also measure the memory footprint of our generative model. We compare it with the current generative rehearsal methods and the space requirements for the memory-buffer based ones. The results of this comparison are shown in Tab.~\ref{tab:results_size}. Our model is much more compact than the ones the other generative rehearsal procedures use. That is thanks to the fact that we have to store only the convolutional decoder of the two parts of the generative model. 
When compared to the rehearsal models with the buffer, our solution requires more space than experience replay methods. However, contrary to such approaches, space requirement for our \ours{} architecture is constant. Our model does not grow with the number of batches served, contrary to the buffer of standard rehearsal methods. While it is not a problem for the simple benchmark datasets with only five batches of data, it becomes crucial when applied in the real-life scenario of continual learning.

\begin{table}[]
  \centering
  \begin{tabular}{l||ccc}
    \toprule
    Model     & MNIST & Fashion & CIFAR--10\\
    & & MNIST & \\ 
    \midrule
    \begin{tabular}{@{}c@{}} Memory-buffer \\ based methods\end{tabular} & 0.7 & 0.7 & 2.9 \\
    \midrule
    AQM~\cite{caccia2020online} & 1.3 & 1.3 & 2.4 \\
    \midrule
    OCDVAE~\cite{2020mundt+4} & 115.9 & 115.9 & - \\
    GEN-MIR~\cite{aljundi2019online} & 5.0 & 11.8 & 34.0 \\
    GR~\cite{2017shin+3} & 4.4 & 15.6 & 32.3 \\
    GR+distill~\cite{vandeven2019scenarios} & 4.4 & 15.6 & 32.3 \\
    RTF~\cite{van2018generative} & 4.4 & 15.6 & 32.3 \\
    {\bf \ours{} (ours)} & 4.6 & 4.6 & 21.0 \\
    \bottomrule
  \end{tabular}
    \caption{Memory requirements of continual learning method (in MB). For generative replay methods, it is calculated as the memory required to store the generative model weights. For memory-buffer based methods, it is the size of the buffer of 100 memories after the last batch. We do not take into account any compression mechanisms neither for images nor for models. However, AQM~\cite{caccia2020online} method includes an embedded compression mechanism and therefore the corresponding results are presented here for the sake of completeness. For generative replay methods (bottom), \ours{} requires comparable (MNIST) or smaller (Fashion MNIST and CIFAR-10) amount of memory comparing to the competing models while yielding higher accuracy, as shown in Tab.~\ref{tab:results_acc}. Moreover, the bigger and more complex the dataset, the more profound the memory savings offered by \ours{} are.}
  \label{tab:results_size}
\end{table}

\subsection{Future work} 
The~\ours{} approach for generative rehearsal presented in this work relies on a binary latent autoencoder and the construction of binary codes living in the latent space. Although the design choices we describe already lead to a significant performance improvement of~\ours{} over the competing methods, we can envision several new research paths that are enabled by our work. The currently used binary encoding parametrization presented in Sec.~\ref{sec:method} relies on a simple yet effective method to populate latent space. One potential direction of future research is the incorporation of other parametrization methods, including trainable models with divergence loss functions. Another path of future work can explore other than binary latent space types and the corresponding encodings. Finally, we believe that the work presented here is a stepping stone towards discovering the relation between generative models used in rehearsal and other compression techniques. Are generative models truly used to compress previously learned knowledge or rather to represent the pattern for generating images of the same kind? Our empirical evaluation indicates that it is rather the latter, yet further research can definitely shed more light on this question.





\section{Conclusions} \label{sec:conclusions} 
In this work, we proposed a novel approach for continual learning with generative replay. Our \ours{} approach uses binary latent space of autoencoder architecture to embed past data batches as binary codes that can be later reconstructed on the fly from a simple and deterministic parametrization function. Inspired by memory-buffer based models, we used binary latent space to store past images as efficient binary codes, while avoiding typical shortcomings of competing generative models that regenerated similar, but lower-quality samples from the past. The evaluation comparison of our approach against state-of-the-art models clearly shows the superiority of \ours{} in terms of average accuracy on three benchmark datasets at a similar or smaller memory footprint than the competing generative rehearsal methods.

{\small
\bibliographystyle{ieee_fullname}
\bibliography{references}

\begin{thebibliography}{10}\itemsep=-1pt

\bibitem{aljundi2019online}
Rahaf Aljundi, Eugene Belilovsky, Tinne Tuytelaars, Laurent Charlin, Massimo
  Caccia, Min Lin, and Lucas Page-Caccia.
\newblock Online {C}ontinual {L}earning with {M}aximally {I}nterfered
  {R}etrieval.
\newblock In {\em NeurIPS}, 2019.

\bibitem{berthelot2018understanding}
David Berthelot, Colin Raffel, Aurko Roy, and Ian Goodfellow.
\newblock Understanding and {I}mproving {I}nterpolation in {A}utoencoders via
  an {A}dversarial {R}egularizer.
\newblock In {\em ICLR}, 2018.

\bibitem{caccia2020online}
Lucas Caccia, Eugene Belilovsky, Massimo Caccia, and Joelle Pineau.
\newblock Online {L}earned {C}ontinual {C}ompression with {A}daptive
  {Q}uantization {M}odules.
\newblock In {\em ICML}, 2020.

\bibitem{carreira15}
M.~Á. {Carreira-Perpiñán} and R. {Raziperchikolaei}.
\newblock Hashing with {B}inary {A}utoencoders.
\newblock In {\em CVPR}, 2015.

\bibitem{2019chaudry+6}
Arslan Chaudhry, Marcus Rohrbach, Mohamed Elhoseiny, Thalaiyasingam Ajanthan,
  Puneet~Kumar Dokania, Philip H.~S. Torr, and Marc'Aurelio Ranzato.
\newblock Continual {L}earning with {T}iny {E}pisodic {M}emories.
\newblock In {\em Multi-Task and Lifelong Reinforcement Learning, Workshop at
  ICML}, 2019.

\bibitem{2019cheung+4}
Brian Cheung, Alexander Terekhov, Yubei Chen, Pulkit Agrawal, and Bruno
  Olshausen.
\newblock Superposition of many models into one.
\newblock In {\em NeurIPS}, 2019.

\bibitem{1999french}
Robert~M. French.
\newblock Catastrophic forgetting in connectionist networks.
\newblock {\em Trends in cognitive sciences}, 1999.

\bibitem{2019golkar+2}
Siavash Golkar, Michael Kagan, and Kyunghyun Cho.
\newblock Continual {L}earning via {N}eural {P}runing.
\newblock In {\em Neuro AI. Workshop at NeurIPS}, 2019.

\bibitem{2014goodfellow+7}
Ian Goodfellow, Jean Pouget-Abadie, Mehdi Mirza, Bing Xu, David Warde-Farley,
  Sherjil Ozair, Aaron Courville, and Yoshua Bengio.
\newblock Generative {A}dversarial {N}etworks.
\newblock In {\em NeurIPS}, 2014.

\bibitem{he17}
Xiangnan He, Lizi Liao, Hanwang Zhang, Liqiang Nie, Xia Hu, and Tat-Seng Chua.
\newblock Neural {C}ollaborative {F}iltering.
\newblock In {\em WWW}, 2017.

\bibitem{2016hendrycks+1}
Dan Hendrycks and Kevin Gimpel.
\newblock A {B}aseline for {D}etecting {M}isclassified and
  {O}ut-of-{D}istribution {E}xamples in {N}eural {N}etworks.
\newblock In {\em ICLR}, 2017.

\bibitem{hinton2015distilling}
Geoffrey Hinton, Oriol Vinyals, and Jeffrey Dean.
\newblock Distilling the {K}nowledge in a {N}eural {N}etwork.
\newblock In {\em Deep Learning and Representation Learning Workshop at
  NeurIPS}, 2015.

\bibitem{kehoe2015survey}
Ben Kehoe, Sachin Patil, Pieter Abbeel, and Ken Goldberg.
\newblock A {S}urvey of {R}esearch on {C}loud {R}obotics and {A}utomation.
\newblock {\em IEEE T-ASE}, 2015.

\bibitem{kingma2014autoencoding}
Diederik~P. Kingma and Max Welling.
\newblock Auto-{E}ncoding {V}ariational {B}ayes.
\newblock In {\em ICLR}, 2014.

\bibitem{2017kirkpatrick+many}
James Kirkpatrick, Razvan Pascanu, Neil Rabinowitz, Joel Veness, Guillaume
  Desjardins, Andrei~A Rusu, Kieran Milan, John Quan, Tiago Ramalho, Agnieszka
  Grabska-Barwinska, Demis Hassabis, Claudia Clopath, Dharshan Kumaran, and
  Raia Hadsell.
\newblock Overcoming catastrophic forgetting in neural networks.
\newblock {\em PNAS}, 2017.

\bibitem{Krizhevsky09learningmultiple}
Alex Krizhevsky.
\newblock Learning {M}ultiple {L}ayers of {F}eatures from {T}iny {I}mages.
\newblock Technical report, 2009.

\bibitem{2020delange+7}
Matthias~De Lange, Rahaf Aljundi, Marc Masana, Sarah Parisot, Xu Jia, Ales
  Leonardis, Gregory Slabaugh, and Tinne Tuytelaars.
\newblock A continual learning survey: {D}efying forgetting in classification
  tasks, 2020.
\newblock arXiv:1909.08383.

\bibitem{lecun1998lenet}
Y. {Lecun}, L. {Bottou}, Y. {Bengio}, and P. {Haffner}.
\newblock Gradient-{B}ased {L}earning {A}pplied to {D}ocument {R}ecognition.
\newblock {\em Proceedings of the IEEE}, 1998.

\bibitem{lecun2010mnist}
Yann LeCun, Corinna Cortes, and CJ Burges.
\newblock {MNIST} handwritten digit database.
\newblock {\em ATT Labs [Online]. Available: http://yann.lecun.com/exdb/mnist},
  2010.

\bibitem{lesort2019generative}
Timoth{\'e}e Lesort, Hugo Caselles-Dupr{\'e}, Michael Garcia-Ortiz, Andrei
  Stoian, and David Filliat.
\newblock Generative {M}odels from the perspective of {C}ontinual {L}earning.
\newblock In {\em IJCNN}, 2019.

\bibitem{lopezpaz2017gradient}
David Lopez-Paz and Marc’Aurelio Ranzato.
\newblock Gradient {E}pisodic {M}emory for {C}ontinual {L}earning.
\newblock In {\em NeurIPS}, 2017.

\bibitem{2018mallya+2}
Arun Mallya, Dillon Davis, and Svetlana Lazebnik.
\newblock Piggyback: {A}dapting a {S}ingle {N}etwork to {M}ultiple {T}asks by
  {L}earning to {M}ask {W}eights.
\newblock In {\em ECCV}, 2018.

\bibitem{2018mallya+1}
Arun Mallya and Svetlana Lazebnik.
\newblock Packnet: {A}dding {M}ultiple {T}asks to a {S}ingle {N}etwork by
  {I}terative {P}runing.
\newblock In {\em CVPR}, 2018.

\bibitem{2018masse+2}
Nicolas~Y. Masse, Gregory~D. Grant, and David~J. Freedman.
\newblock Alleviating catastrophic forgetting using context-dependent gating
  and synaptic stabilization.
\newblock {\em PNAS}, 2018.

\bibitem{mena19}
Francisco Mena and Ricardo Nanculef.
\newblock A {B}inary {V}ariational {A}utoencoder for {H}ashing.
\newblock In {\em CIARP}, 2019.

\bibitem{2020mundt+3}
Martin Mundt, Yong~Won Hong, Iuliia Pliushch, and Visvanathan Ramesh.
\newblock A {W}holistic {V}iew of {C}ontinual {L}earning with {D}eep {N}eural
  {N}etworks: {F}orgotten {L}essons and the {B}ridge to {A}ctive and {O}pen
  {W}orld {L}earning, 2020.
\newblock arXiv:2009.01797.

\bibitem{2020mundt+4}
Martin Mundt, Sagnik Majumder, Iuliia Pliushch, Yong~Won Hong, and Visvanathan
  Ramesh.
\newblock Unified {P}robabilistic {D}eep {C}ontinual {L}earning through
  {G}enerative {R}eplay and {O}pen {S}et {R}ecognition, 2020.
\newblock arXiv:1905.12019v4.

\bibitem{rebuffi2017icarl}
S. {Rebuffi}, A. {Kolesnikov}, G. {Sperl}, and C.~H. {Lampert}.
\newblock i{C}a{RL}: Incremental {C}lassifier and {R}epresentation {L}earning.
\newblock In {\em CVPR}, 2017.

\bibitem{2019rolnick+4}
David Rolnick, Arun Ahuja, Jonathan Schwarz, Timothy Lillicrap, and Gregory
  Wayne.
\newblock Experience {R}eplay for {C}ontinual {L}earning.
\newblock In {\em NeurIPS}, 2019.

\bibitem{2016rusu+7}
Andrei~A. Rusu, Neil~C. Rabinowitz, Guillaume Desjardins, Hubert Soyer, James
  Kirkpatrick, Koray Kavukcuoglu, Razvan Pascanu, and Raia Hadsell.
\newblock Progressive {N}eural {N}etworks, 2016.
\newblock arXiv:1606.04671.

\bibitem{2017shin+3}
Hanul Shin, Jung~Kwon Lee, Jaehong Kim, and Jiwon Kim.
\newblock Continual {L}earning with {D}eep {G}enerative {R}eplay.
\newblock In {\em NeurIPS}, 2017.

\bibitem{tilaro2018model}
Filippo Tilaro, Benjamin Bradu, Manuel Gonzalez-Berges, Mikhail Roshchin, and
  Fernando Varela.
\newblock Model {L}earning {A}lgorithms for {A}nomaly {D}etection in {CERN}
  {C}ontrol {S}ystems.
\newblock In {\em ICALEPCS}, 2018.

\bibitem{van2018generative}
Gido~M van~de Ven and Andreas~S Tolias.
\newblock Generative replay with feedback connections as a general strategy for
  continual learning, 2018.
\newblock arXiv:1809.10635.

\bibitem{vandeven2019scenarios}
Gido~M. van~de Ven and Andreas~S. Tolias.
\newblock Three scenarios for continual learning, 2019.
\newblock arXiv:1904.07734.

\bibitem{oord2018neural}
Aaron van~den Oord, Oriol Vinyals, and Koray Kavukcuoglu.
\newblock Neural {D}iscrete {R}epresentation {L}earning.
\newblock In {\em NeurIPS}, 2017.

\bibitem{2020wen+2}
Yeming Wen, Dustin Tran, and Jimmy Ba.
\newblock Batch{E}nsemble: an {A}lternative {A}pproach to {E}fficient
  {E}nsemble and {L}ifelong {L}earning.
\newblock In {\em ICLR}, 2020.

\bibitem{2020wortsman+6}
Mitchell Wortsman, Vivek Ramanujan, Rosanne Liu, Aniruddha Kembhavi, Mohammad
  Rastegari, Jason Yosinski, and Ali Farhadi.
\newblock Supermasks in {S}uperposition.
\newblock In {\em arXiv:2006.14769, accepted for publication in NeurIPS}, 2020.

\bibitem{2019xiang+3}
Ye Xiang, Ying Fu, Pan Ji, and Hua Huang.
\newblock Incremental {L}earning {U}sing {C}onditional {A}dversarial
  {N}etworks.
\newblock In {\em ICCV}, 2019.

\bibitem{xiao2017fashionmnist}
Han Xiao, Kashif Rasul, and Roland Vollgraf.
\newblock Fashion-{MNIST}: a {N}ovel {I}mage {D}ataset for {B}enchmarking
  {M}achine {L}earning {A}lgorithms, 2017.
\newblock arXiv:1708.07747.

\bibitem{2018xu+1}
Ju Xu and Zhanxing Zhu.
\newblock Reinforced {C}ontinual {L}earning.
\newblock In {\em NeurIPS}, 2018.

\bibitem{yang2018pixor}
Bin Yang, Wenjie Luo, and Raquel Urtasun.
\newblock Pixor: {R}eal-time 3d {O}bject {D}etection from {P}oint {C}louds.
\newblock In {\em CVPR}, 2018.

\bibitem{2017yoon+3}
Jaehong Yoon, Eunho Yang, Jeongtae Lee, and Sung~Ju Hwang.
\newblock Lifelong {L}earning with {D}ynamically {E}xpandable {N}etworks.
\newblock In {\em ICLR}, 2018.

\bibitem{zamorski20}
Maciej Zamorski, Maciej Zieba, Piotr Klukowski, Rafal Nowak, Karol Kurach,
  Wojciech Stokowiec, and Tomasz Trzcinski.
\newblock Adversarial {A}utoencoders for {C}ompact {R}epresentations of 3{D}
  {P}oint {C}louds.
\newblock {\em Comput. Vis. Image Underst.}, 2020.

\bibitem{2017zenke+2}
Friedemann Zenke, Ben Poole, and Surya Ganguli.
\newblock Continual {L}earning {T}hrough {S}ynaptic {I}ntelligence.
\newblock In {\em ICML}, 2017.

\end{thebibliography}
}

\section*{Appendix}
\section*{Proof of Lemma 1} \label{sec:proof} 

We have 
$$
    p^{\lfloor m \ln 2 / \ln p \rfloor} 
    < p^{m \ln 2 / \ln p} 
    = p^{(\log_p 2)m} 
    = 2^m,  
$$
and
$$
    p^{\lfloor m \ln 2 / \ln p \rfloor}  
    > p^{m \ln 2 / \ln p - 1}  
    = p^{(\log_p 2)m - 1} 
    = 2^m/p 
$$
which proves point 1. 

In order to prove point 2, we notice that since $p$ is prime, and $p\neq2$, $p$ raised to any natural power is coprime with $2^m$. Suppose there are two different $i,j \in \{1, \dots, 2^m\}$ that have the same code in the form of $m$ least significant bits of (1). That would mean that 
$$
    |i-j|*p^{\lfloor m \ln 2 / \ln p \rfloor}\!\!\!\!\mod 2^m = 0. 
$$
That can not be true since $p^{\lfloor m \ln 2 / \ln p \rfloor}$ has no common divisors with $2^m$ and $|i-j|$ is not divisible by $2^m$ because $|i-j|<2^m$. \qed

\section*{Models architectures}

\subsection*{MNIST and FashionMNIST}
For the encoder, we use three convolutional layers with 32, 64, and 128 filters, respectively, with kernel size 3×3 and 2×2 stride. We compress the resulting feature map to the vector of 2048 values, which we translate to the binary latent space of 200 neurons. For the decoder we map features from the latent space through two fully connected layers with 800 neurons, then it is forwarded through 3 transposed convolutional layers with 32, 128, 192 filters with 3×3 kernel size, 2×2 stride, and the final output layer with 4×4 kernel size with no stride. For all activation functions, we use LeakyRelu. We also employ batch normalization after each convolutional layer.

Our classifier is based on the LeNet architecture with 3 convolutional layers with 32, 64, and 128 filters with 4×4 kernel size. We cast the output of the last convolution into the feature space of 1152 values, which we pass through the fully connected layer of 84 neurons to form the final output. We use an additional batch normalization after each convolutional layer, and dropout to prevent the network from overfitting.

\subsection*{CIFAR-10}
For the CIFAR-10 dataset, we employ similar network architectures as for MNIST and Fashion-MNIST but with a greater number of filters and bigger latent size. 

For the encoder, we use 3 convolutional layers with 32, 64, and 128 filers of 3×3 kernel size and 2×2 stride. After that, we encode the resulting feature map of 2048 features into the latent space of 1000 neurons for the index encoding and 10 neurons for the batch number encodings. 
In the decoder, similarly to the previous one, we use two fully connected layers to gradually map 1010 binary values from the latent space through the fully connected layer of 1500 neurons to the feature map of 6144 neurons. 

We translate encoded features through 3 transposed convolution layers with 512, 768, and 640 features, 3×3, 4×4, 4×4 kernel size, and 2×2, 2×2, and 1×1 stride. The final transposed convolution layer translates filters to the final 3 channels with 4×4 kernel and no stride.

\section*{Visualization of generated samples}

To emphasize the quality of generated samples we show in higher resolution, the visualization of CIFAR-10 images generation with our BinPlay Binary Latent Autoencdeor after each batch. We visualize the same instances to show noticeable, but slow process of quality degradation. 

\begin{figure*}
\centering
\subfigure[Batch one]{\includegraphics[width=.49\linewidth]{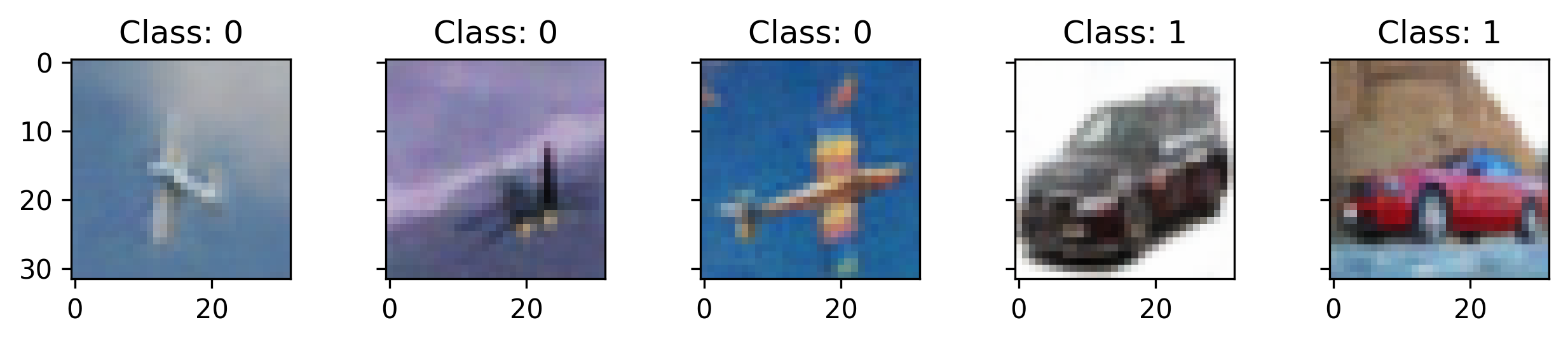}}
\subfigure[Batch two]{\includegraphics[width=.49\linewidth]{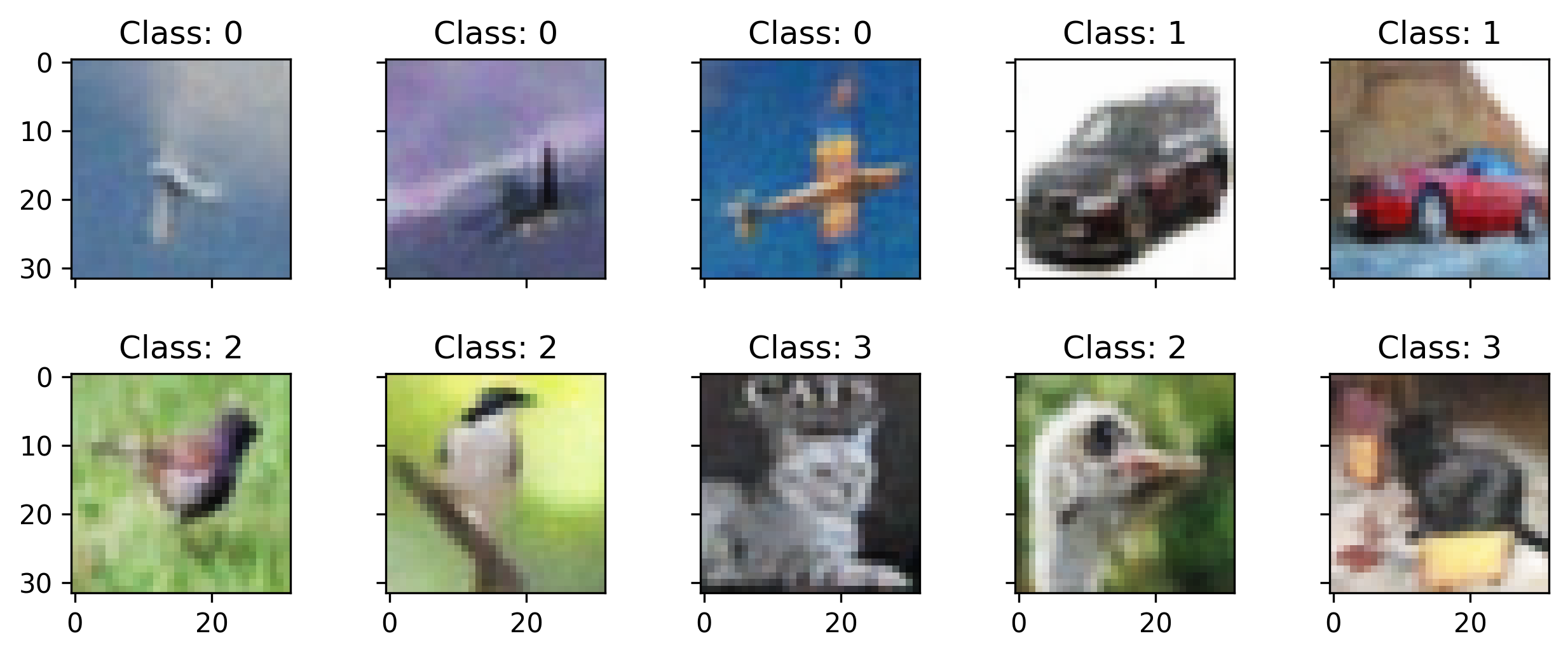}}
\subfigure[Batch three]{\includegraphics[width=.49\linewidth]{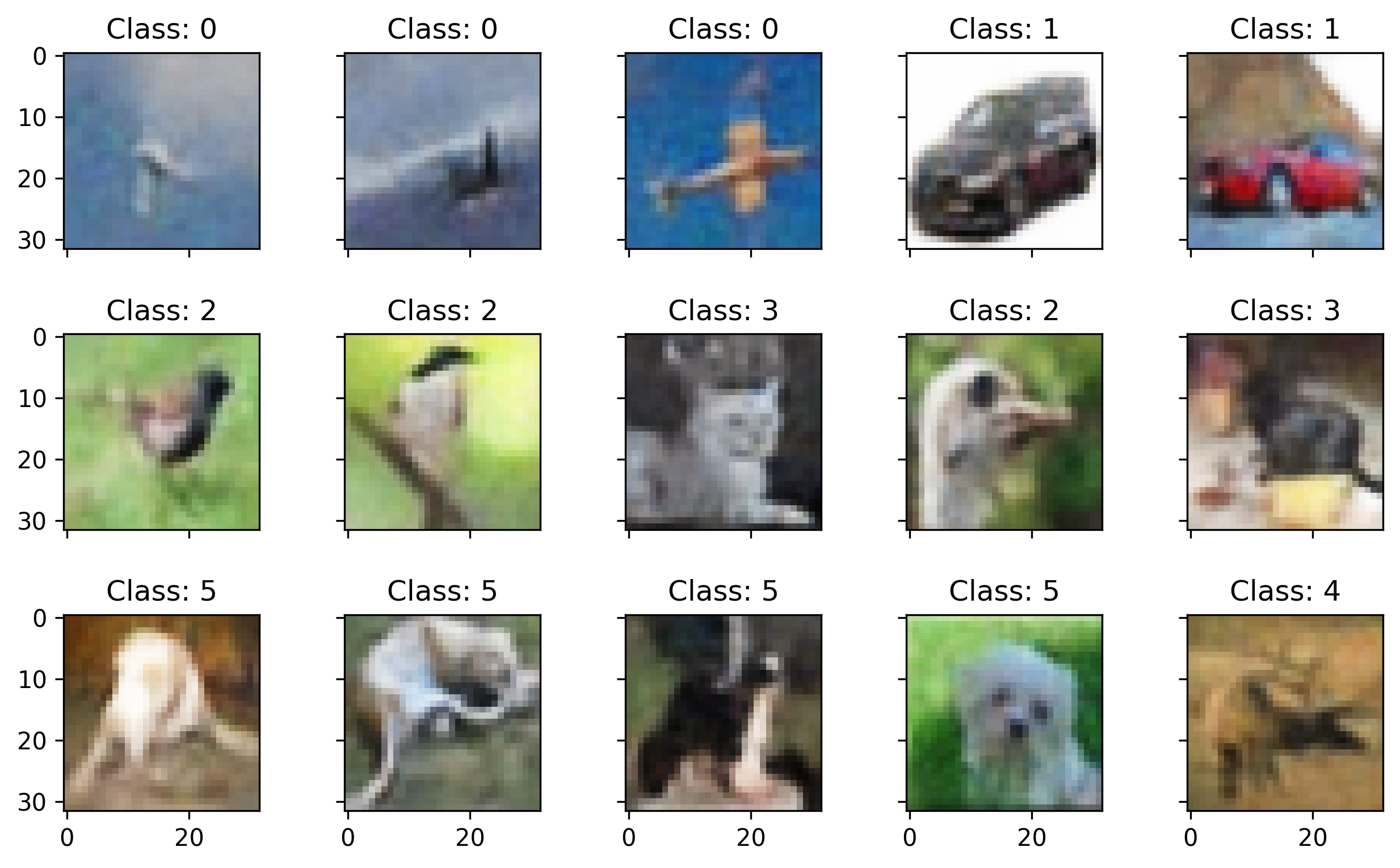}}
\subfigure[Batch four]{\includegraphics[width=.49\linewidth]{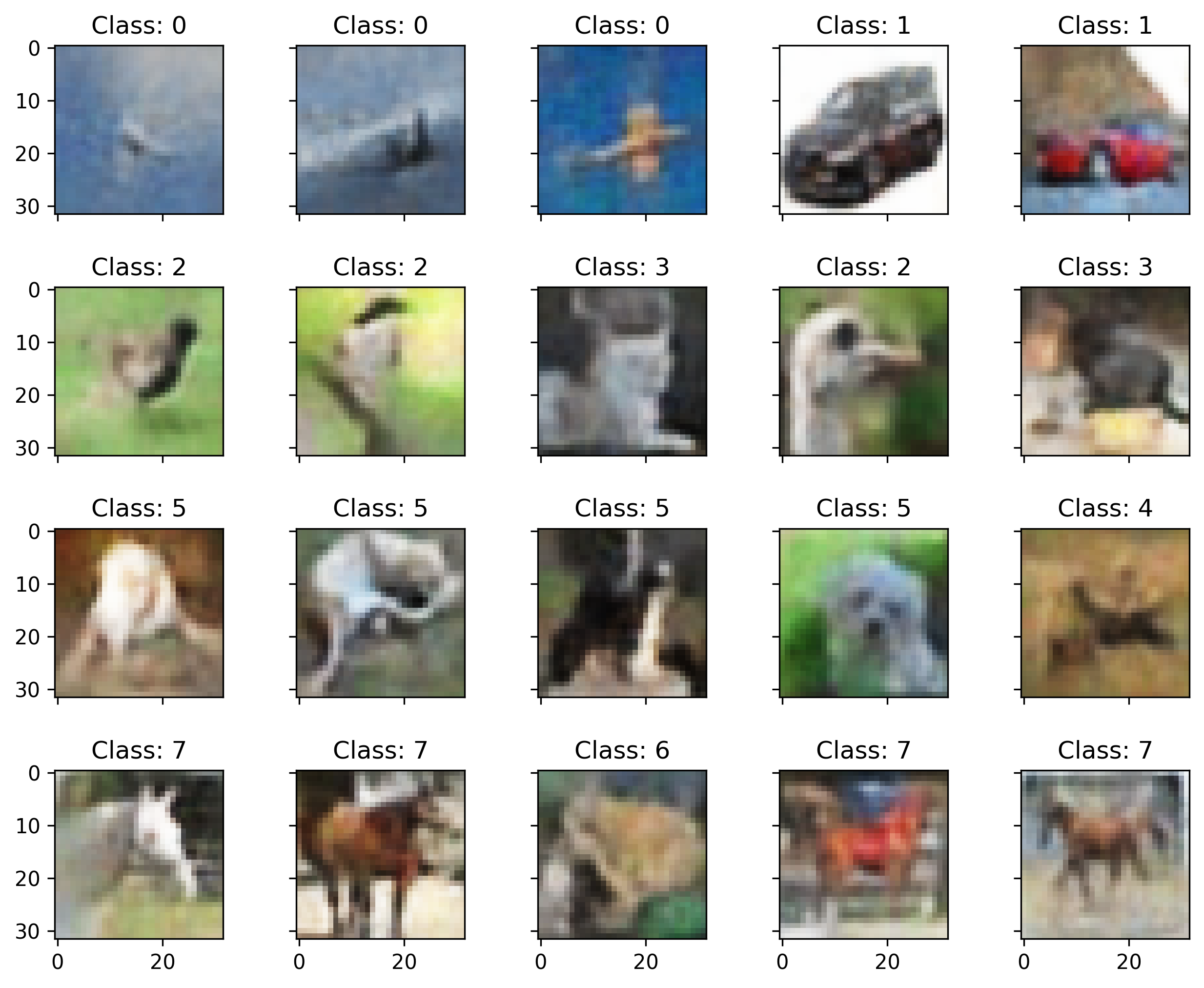}}
\subfigure[Batch five]{\includegraphics[width=.49\linewidth]{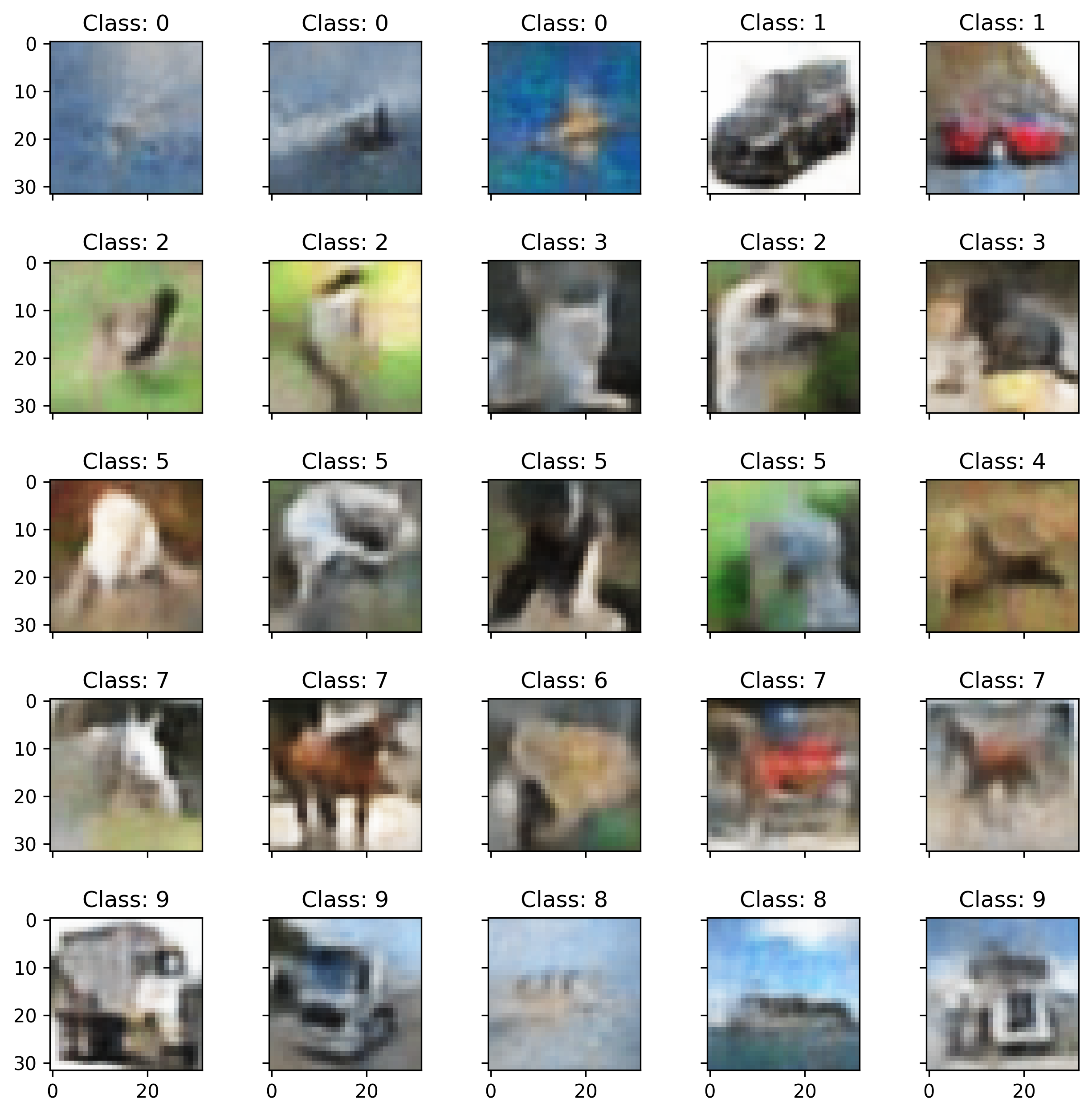}}

\caption{Images generated by our BinPlay binary latent autoencoder after consecutive number of batches}
\label{fig:generations_full} 
\end{figure*} 

\end{document}